# Learning Visual N-Grams from Web Data

Ang Li*
University of Maryland
College Park, MD 20742, USA
angli@umiacs.umd.edu

Allan Jabri    Armand Joulin    Laurens van der Maaten
Facebook AI Research
770 Broadway, New York, NY 10025, USA
{ajabri,ajoulin,lvdmaaten}@fb.com

## Abstract

*Real-world image recognition systems need to recognize tens of thousands of classes that constitute a plethora of visual concepts. The traditional approach of annotating thousands of images per class for training is infeasible in such a scenario, prompting the use of webly supervised data. This paper explores the training of image-recognition systems on large numbers of images and associated user comments,* without *using manually labeled images. In particular, we develop visual $n$-gram models that can predict arbitrary phrases that are relevant to the content of an image. Our visual $n$-gram models are feed-forward convolutional networks trained using new loss functions that are inspired by $n$-gram models commonly used in language modeling. We demonstrate the merits of our models in phrase prediction, phrase-based image retrieval, relating images and captions, and zero-shot transfer.*

## 1. Introduction

Research on visual recognition models has traditionally focused on supervised learning models that consider only a small set of discrete classes, and that learn their parameters from datasets in which (1) all images are manually annotated for each of these classes and (2) a substantial number of annotated images is available to define each of the classes. This tradition dates back to early image-recognition benchmarks such as CalTech-101 [18] but is still common in modern benchmarks such as ImageNet [47] and COCO [42]. The assumptions that are implicit in such benchmarks are at odds with many real-world applications of image-recognition systems, which often need to be deployed in an *open-world* setting [3]. In the open-world setting, the number of classes to recognize is potentially very large and class types are wildly varying [13]: they include generic objects such as "dog" or "car", landmarks such as "Golden Gate Bridge" or "Times Square", scenes such as "city park" or "street market", and actions such as "speed walking" or "public speaking". The traditional approach of manually annotating images for training does not scale well to the open-world setting because of the amount of effort required to gather and annotate images for all relevant classes. To circumvent this problem, several recent studies have tried to use image data from photo-sharing websites such as Flickr to train their models [5, 9, 19, 28, 41, 46, 57, 58, 62]: such images have no manually curated annotations, but they do have metadata such as tags, captions, comments, and geo-locations that provide weak information about the image content, and are readily available in nearly infinite numbers.

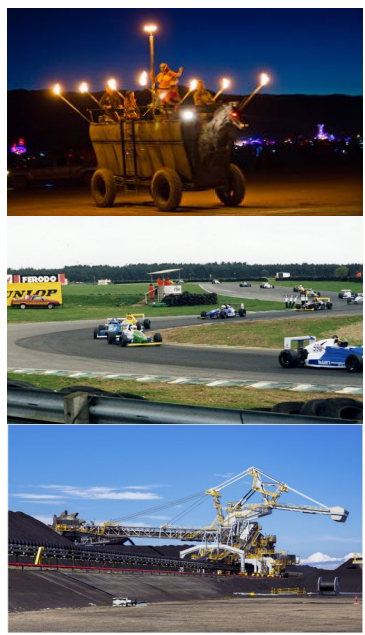

Figure 1. Four high-scoring visual $n$-grams for three images in our test set according to our visual $n$-gram model, which was trained *solely* on *unsupervised* web data. We selected the $n$-grams that are displayed in the figure from the five highest scoring $n$-grams according to our model, in such a way as to minimize word overlap between the $n$-grams. For all figures in the paper, we refer the reader to the supplementary material for license information.

---

*This work was done while Ang Li was at Facebook AI Research.

In this paper, we follow [28] and study the training of models on images and their associated user comments present in the YFCC100M dataset [55]. In particular, we aim to take a step in bridging the semantic gap between vision and language by *predicting phrases that are relevant to the contents of an image*. We develop *visual n-gram* models that, given an image **I**, assign a likelihood $p(w|\mathbf{I})$ to each possible phrase ($n$-gram) $w$. Our models are convolutional networks trained using a loss function that is motivated by $n$-gram smoothers commonly used in language modeling [26, 34]: we develop a novel, differentiable loss function that optimizes trainable parameters for frequent $n$-grams, whereas for infrequent $n$-grams, the loss is dominated by the predicted likelihood of smaller "sub-grams". The resulting visual $n$-gram models have substantial advantages over prior open-world visual models [28]: they recognize landmarks such as "Times Square", they differentiate between 'Washington DC' and the "Washington Nationals", and they distinguish between "city park" and "Park City".

The technical contributions of this paper are threefold: (1) we are the first to explore the prediction of $n$-grams relevant to image content using convolutional networks, (2) we develop a novel, differentiable smoothing layer for such networks, and (3) we provide a simple solution to the out-of-vocabulary problem of traditional image-recognition models. We present a series of experiments to demonstrate the merits of our proposed model in image tagging, image retrieval, image captioning, and zero-shot transfer.

## 2. Related Work

There is a substantial body of prior work that is related to this study, in particular, work on (1) learning from weakly supervised web data, (2) relating image content and language, and (3) language modeling. We give a (non-exhaustive) overview of prior work below.

**Learning from weakly supervised web data.** Several prior studies have used Google Images to obtain large collections of (weakly) labeled images for the training of vision models [5, 9, 19, 46, 54, 58, 62]. We do not opt for such an approach here because it is very difficult to understand the biases it introduces, in particular, because image retrieval by Google Images is likely aided by a content-based image retrieval model itself. This introduces the real danger that training on data from Google Images amounts to replicating an existing black-box vision system. Various other studies have used data from photo-sharing websites such as Flickr for training; for instance, to train hierarchical topic models [38] or multiple-instance learning SVMs [39], to learn label distribution models [12, 64], to finetune pretrained convolutional networks [24], and to train weak classifiers that produce additional visual features [56]. Like this study, [28] trains convolutional networks on the image-comment pairs. Our study differs[1] from [28] in that we do not just consider single words, as a result of which our models distinguish between, *e.g.*, "city park" and "Park City".

**Relating image content and language.** Our approach is connected to a wide body of work that aims at bridging the semantic gap between vision and language [49]. In particular, many studies have explored this problem in the context of image captioning. Most image-captioning systems train a recurrent network or maximum entropy language model on top of object classifications produced by a convolutional network; the models are either trained separately [14, 29, 43] or end-to-end [15, 59]. We do not consider recurrent networks in our study because test-time inference in such networks is slow, which hampers the deployment of such models in real-world applications. An image-captioning study that is closely related to our work is [37], which trains a bilinear model that outputs phrase probabilities given an image feature and combines the relevant phrases into a caption using a collection of heuristics. Several other works have explored joint embedding of images and text, either at the word level [20] or at the sentence level [17, 30]. What distinguishes our study is that prior work is generally limited in the variety of visual concepts it can deal with; these studies rely on vision models that recognize only small numbers of classes and / or on the availability of "ground-truth" captions that describe the image content — such captions are very different from a typical user comment on Flickr. In contrast to prior work, we consider the open-world setting with very large numbers of visual concepts, and we do not rely on ground-truth captions provided by human annotators. Our study is most similar to that of [40], which uses $n$-gram to generate image descriptions; unlike [40], we we do not rely on separately trained image-classification pipelines. Instead, we train our model end-to-end on a dataset without ground-truth labels.

**Language models.** Several prior studies have used phrase embeddings for natural language processing tasks such as named entity recognition [45], text classification [27, 53, 61], and machine translation [68, 71]. These studies differ from our work in that they focus solely on language modeling and not on visual recognition. Our models are inspired by smoothing techniques used in traditional $n$-gram language models[2], in particular, Jelinek-Mercer smoothing [26]. Our models differ from traditional $n$-gram language models in that they are *image-conditioned* and *parametric*: whereas $n$-gram models count the frequency of $n$-grams in a text corpus to produce a distribution over phrases or sentences, our model measures phrase likelihoods by evaluating inner products between image features and learned parameter vectors.

---

[1] Indeed, the models in [28] are a special case of our models in which only unigrams are considered.
[2] A good overview of these techniques is given in [8, 22].

## 3. Learning Visual N-Gram Models

Below, we describe the dataset we use in our experiments, the loss functions we optimize, and the training procedure we use for optimization.

### 3.1. Dataset

We train our models on the YFCC100M dataset, which contains 99.2 million images and associated multi-lingual user comments [55]. We applied a simple language detector to the dataset to select only images with English user comments, leaving a total of 30 million examples for training and testing. We preprocessed the text by removing punctuations, and we added [BEGIN] and [END] tokens at the beginning and end of each sentence. We preprocess all images by rescaling them to $256 \times 256$ pixels (using bicubic interpolation), cropping the central $224 \times 224$, subtracting the mean pixel value of each image, and dividing by the standard deviation of the pixel values.

For most experiments, we use a dictionary of all English $n$-grams (with $n$ between 1 and 5) with more than $1,000$ occurrences in the 30 million English comments. This dictionary contains $142,806$ $n$-grams: $22,869$ unigrams, $56,830$ bigrams, $32,560$ trigrams, $17,351$ four-grams, and $13,196$ five-grams. We emphasize that the smoothed visual $n$-gram models we describe below are trained and evaluated on all $n$-grams in the dataset, even if these $n$-grams are not in the dictionary. However, whereas the probability of in-dictionary $n$-grams is primarily a function of parameters that are specifically tuned for those $n$-grams, the probability of out-of-dictionary $n$-grams is composed from the probability of smaller in-dictionary $n$-grams (details below).

### 3.2. Loss functions

The main contribution of this paper is in the loss functions we use to train our phrase prediction models. In particular, we explore (1) a *naive n-gram loss* that measures the (negative) log-likelihood of in-dictionary $n$-grams that are present in a comment and (2) a *smoothed n-gram loss* that measures the (negative) log-likelihood of all $n$-grams, even if these $n$-grams are not in the dictionary. This loss uses smoothing to assign non-zero probabilities to out-of-dictionary $n$-grams; specifically, we experiment with Jelinek-Mercer smoothing [26].

**Notation.** We denote the input image by $\mathbf{I}$ and the image features extracted by the convolutional network with parameters $\theta$ by $\phi(\mathbf{I}; \theta) \in \mathbb{R}^D$. We denote the $n$-gram dictionary that our model uses by $\mathcal{D}$ and a comment containing $K$ words by $w \in [1, C]^K$, where $C$ is the total number of words in the (English) language. We denote the $n$-gram that ends at the $i$-th word of comment $w$ by $w_{i-n+1}^i$ and the $i$-th word in comment $w$ by $w_i^i$. Our predictive distribution is governed by a $n$-gram embedding matrix $\mathbf{E} \in \mathbb{R}^{D \times |\mathcal{D}|}$. With a slight abuse of notation, we denote the embedding corresponding to a particular $n$-gram $w$ by $\mathbf{e}_w$. For brevity, we omit the sum over all image-comment pairs in the training / test data when writing loss functions.

**Naive $n$-gram loss.** The naive $n$-gram loss is a standard multi-class logistic loss over all $n$-grams in the dictionary $\mathcal{D}$. The loss is summed over all $n$-grams that appear in the sentence $w$; that is, $n$-grams that do not appear in the dictionary are ignored:

$$\ell(\mathbf{I}, w; \theta, \mathbf{E}) = -\sum_{m=1}^{n}\sum_{i=n}^{K} \mathbb{I}\left[w_{i-m+1}^i \in \mathcal{D}\right]$$
$$\log p_{obs}\left(w_{i-m+1}^i | \phi(\mathbf{I}; \theta); \mathbf{E}\right),$$

where the *observational likelihood* $p_{obs}(\cdot)$ is given by a softmax distribution over all in-dictionary $n$-grams $w$ that is governed by the inner product between the image features $\phi(\mathbf{I}; \theta)$ and the $n$-gram embeddings:

$$p_{obs}\left(w | \phi(\mathbf{I}; \theta); \mathbf{E}\right) = \frac{\exp\left(-\mathbf{e}_w^\top \phi(\mathbf{I}; \theta)\right)}{\sum_{w' \in \mathcal{D}} \exp\left(-\mathbf{e}_{w'}^\top \phi(\mathbf{I}; \theta)\right)}.$$

The image features $\phi(\mathbf{I}; \theta)$ are produced by a convolutional network $\phi(\cdot)$, which we describe in more detail in 3.3.

The naive $n$-gram loss cannot do language modeling because it does not model a conditional probability. To circumvent this issue, we construct an ad-hoc conditional distribution based on the scores produced by our model at prediction time using a "stupid" back-off model [6]:

$$p\left(w_i^i | w_{i-n+1}^{i-1}\right) \propto \begin{cases} p_{obs}\left(w_i^i | w_{i-n+1}^{i-1}\right), & \text{if } w_{i-n+1}^i \in \mathcal{D} \\ \lambda p\left(w_i^i | w_{i-n+2}^{i-1}\right), & \text{otherwise.} \end{cases}$$

For brevity, we dropped the conditioning on $\phi(\mathbf{I}; \theta)$ and $\mathbf{E}$.

**Jelinek-Mercer (J-M) loss.** The simple $n$-gram loss has two main disadvantages: (1) it ignores out-of-dictionary $n$-grams entirely during training and (2) the parameters $\mathbf{E}$ that correspond to infrequent in-dictionary words are difficult to pin down. Inspired by Jelinek-Mercer smoothing, we propose a loss function that aims to address both these issues:

$$\ell(\mathbf{I}, w; \theta, \mathbf{E}) = -\sum_{i=1}^{K} \log p\left(w_i^i | w_{i-n+1}^{i-1}, \phi(\mathbf{I}; \theta); \mathbf{E}\right),$$

where the likelihood of a word conditioned on the $(n-1)$ words appearing before it is defined as:

$$p\left(w_i^i | w_{i-n+1}^{i-1}\right) = \lambda p_{obs}\left(w_i^i | w_{i-n+1}^{i-1}\right) + (1-\lambda) p\left(w_i^i | w_{i-n+2}^{i-1}\right).$$

Herein, we removed the conditioning on $\phi(\mathbf{I}; \theta)$ and $\mathbf{E}$ for brevity. The parameter $0 \leq \lambda \leq 1$ is a smoothing constant that governs how much of the probability mass from $(n-1)$-grams is (recursively) transferred to both in-dictionary and out-of-dictionary $n$-grams. The probability mass transfer prevents the Jelinek-Mercer loss from assigning zero probability (which would lead to infinite loss) to out-of-vocabulary $n$-grams, and it allows it to learn from low-frequency and out-of-vocabulary $n$-grams.

The Jelinek-Mercer loss proposed above is different from traditional is Jelinek-Mercer smoothing: in particular, it is *differentiable* with respect to both $\mathbf{E}$ and $\theta$. As a result, the loss can be backpropagated through the convolutional network. In particular, the loss gradient with respect to $\phi$ is given by:

$$\frac{\partial \ell}{\partial \phi} = - \sum_{i=1}^{K} p\left(w_i^i | w_{i-n+1}^{i-1}, \phi(\mathbf{I}; \theta); \mathbf{E}\right) \frac{\partial p}{\partial \phi},$$

where the partial derivatives are given by:

$$\frac{\partial p}{\partial \phi} = \lambda \frac{\partial p_{obs}}{\partial \phi} + (1-\lambda) \frac{\partial p}{\partial \phi}$$

$$\frac{\partial p_{obs}}{\partial \phi} = p_{obs}(w|\phi(\mathbf{I}; \theta); \mathbf{E}) \left(\mathbb{E}[\mathbf{e}_{w'}]_{w' \sim p_{obs}} - \mathbf{e}_w\right).$$

This error signal can be backpropagated directly through the convolutional network $\phi(\cdot)$.

### 3.3. Training

The core of our visual recognition models is formed by a convolutional network $\phi(\mathbf{I}; \theta)$. For expediency, we opt for a residual network [23] with 34 layers. Our networks are initialized by an Imagenet-trained network, and trained to minimize the loss functions described above using stochastic gradient descent using a batch size of 128 for 10 epochs. In all experiments, we employ Nesterov momentum of 0.9, a weight decay of 0.0001, and an initial learning rate of 0.1; the learning rate is divided by 10 whenever the training loss stabilizes (until a minimum learning rate of 0.001).

A major bottleneck in training is the large number of outputs of our observation model: doing a forward-backward pass with 512 inputs (the image features) and $142,806$ outputs (the $n$-grams) is computationally intensive. To circumvent this issue, we follow [28] and perform *stochastic gradient descent over outputs* [4]: we only perform the forward-backward pass for a random subset (formed by all positive $n$-grams in the batch) of the columns of $\mathbf{E}$. This simple approximation works well in practice, and it can be shown to be closely related to the exact loss [28].

## 4. Experiments

Below, we present the four sets of experiments we performed to assess the performance of our visual $n$-gram models in: (1) phrase-level image tagging, (2) phrase-based image retrieval, (3) relating images and captions, and (4) zero-shot transfer.

| Loss / Smoothing | "Stupid" back-off | Jelinek-Mercer |
|---|---|---|
| Imagenet + linear | 349 | 233 |
| Naive $n$-gram | 297 | 212 |
| Jelinek-Mercer | 276 | **199** |

Table 1. Perplexity of visual $n$-gram models averaged over YFCC100M test set of $10,000$ images (evaluated on in-dictionary words only). Results for two losses (rows) with and without smoothing at test time (columns). Lower is better.

### 4.1. Phrase-level image tagging

We first gauge whether relevant comments for images have high likelihood under our visual $n$-gram models. Specifically, we measure the perplexity of predicting the correct words in a comment on a held-out test set of $10,000$ images, and average this perplexity over all images in the test set. The perplexity of a model is defined as $2^{H(p)}$, where $H(p)$ is the cross-entropy:

$$H(p) = -\frac{1}{K} \sum_{i=1}^{K} \log_2 p\left(w_i^i | w_{i-n+1}^{i-1}, \phi(\mathbf{I}; \theta); \mathbf{E}\right).$$

We only consider in-dictionary unigrams in our perplexity measurements. As is common in language modeling [22], we assume a uniform conditional distribution $p_{obs}\left(w_i^i | w_{i-n+1}^{i-1}\right)$ for $n$-grams whose prefix is not in the dictionary (*i.e.*, for $n$-grams for which $w_{i-n+1}^{i-1} \notin \mathcal{D}$). Based on the results of preliminary experiments on a held-out validation set, we set $\lambda = 0.2$ in the Jelinek-Mercer loss.

We compare models that use either of the two loss functions (the naive in-dictionary $n$-gram loss and Jelinek-Mercer loss) with a baseline trained with a linear layer on top of Imagenet-trained visual features trained using naive $n$-gram loss. We consider two settings of our models at prediction time: (1) a setting in which we use the "stupid" back-off model with $\lambda = 0.6$; and (2) a setting in which we smooth the $p(\cdot)$ predictions using Jelinek-Mercer smoothing (as described above) using $\lambda = 0.2$.

The resulting perplexities for all experimental settings are presented in Table 1. From the results presented in the table, we observe that: (1) the use of smoothing losses for training image-based phrase prediction models leads to better models than the use of a naive $n$-gram loss; and (2) the use of additional smoothing at test time may further reduce the perplexity of the $n$-gram model. The former effect is the result of the ability of smoothing losses to direct the learning signal to the most relevant $n$-grams instead of equally spreading it over all $n$-grams that are present in the target. The latter effect is the result of the ability of prediction-time smoothing to propagate the probability mass from in-dictionary $n$-grams to relevant out-of-dictionary $n$-grams.

To obtain more insight into the phrase-prediction performance of our models, we also assess our model's ability to predict relevant phrases ($n$-grams) for images. To correct for variations in the marginal frequency of $n$-grams, we calibrate all log-likelihood scores by subtracting the average log-likelihood our model predicts on a large collection of held-out validation images. We predict $n$-gram phrases for images by outputting the $n$-grams with the highest calibrated log-likelihood score for an image. Examples of the resulting $n$-gram predictions are shown in Figure 1.

| Model | R@1 | R@5 | R@10 | Accuracy |
|---|---|---|---|---|
| Imagenet + linear | 5.0 | 10.7 | 14.5 | 32.7 |
| Naive $n$-gram | 5.5 | 11.6 | 15.1 | 36.4 |
| Jelinek-Mercer | **6.2** | **13.0** | **18.1** | **42.0** |

Table 2. Phrase-prediction performance on YFCC100M test set of $10,000$ images measured in terms of recall@$k$ at three cut-off levels $k$ (lefthand-side; see text for details) and the percentage of correctly predicted $n$-grams according to human raters (righthand-side) for one baseline model and two of our phrase prediction models. Higher is better.

We quantify phrase-prediction performance in terms of recall@$k$ on a set of $10,000$ images from the YFCC100M test set. We define recall@$k$ as the average percentage of $n$-grams appearing in the comment that are among the $k$ front-ranked $n$-grams when the $n$-grams are sorted according to their score under the model. In this experiment and all experiments hereafter, we only present results where the same smoothing is used at training and at prediction time: that is, we use the "stupid" back-off model on the predictions of naive $n$-grams models and we smooth the predictions of Jelinek-Mercer models using Jelinek-Mercer smoothing. As a baseline, we consider a linear multi-class classifier over $n$-grams (*i.e.*, using naive $n$-gram loss) trained on features produced by an Imagenet-trained convolutional network. The results are shown in the lefthand-side of Table 2.

Because the $n$-grams in the YFCC100M test set are noisy targets (many words that are relevant to the image content are not present in the comments), we also performed an experiment on Amazon Mechanical Turk in which we asked two human raters whether or not the highest-scoring $n$-gram was relevant to the content of the image. We filter out unreliable raters based on their response time, and for each of our models, we measure the percentage of retrieved $n$-grams that is considered relevant by the remaining raters. The resulting accuracies of the visual $n$-gram models are reported in the righthand-side of Table 2.

The results presented in the table are in line with the results presented in Table 1: they show that the use of a smoothing loss substantially improves the results compared to baseline models based on the naive $n$-gram loss. In particular, the relative performance in recall@$k$ between our best model and the Imagenet-trained baseline model is approximately $20\%$. The merits of the Jelinek-Mercer loss are confirmed by our experiment on Mechanical Turk: according to human annotators, $42.0\%$ of the predicted phrases is relevant to the visual content of the image.

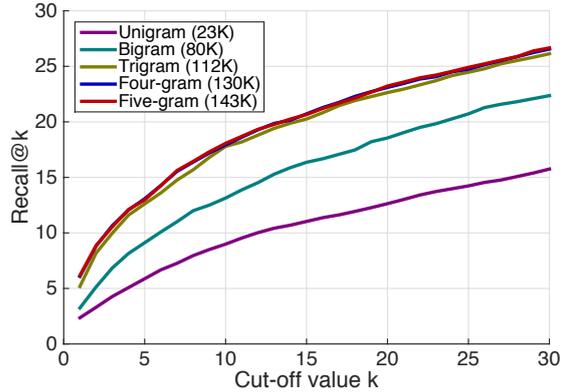

Figure 2. Recall@$k$ on $n$-gram retrieval of five models with increasing maximum length of $n$-grams included in the dictionary ($n=1,\ldots,5$), for varying cut-off values $k$. The dictionary size of each of the models is shown between brackets. Higher is better.

Next, we study the performance of our Jelinek-Mercer model as a function of $n$; that is, we investigate the effect of including longer $n$-grams in our model on the model performance. As before, we measure recall@$k$ of $n$-gram retrieval as a function of the cut-off level $k$, and consider models with unigrams to five-grams. Figure 2 presents the results of this experiment, which shows that the performance of our models increases as we include longer $n$-grams in the dictionary. The figure also reveals diminishing returns: the improvements obtained from going beyond trigrams are limited.

### 4.2. Phrase-based image retrieval

In the second set of experiments, we measure the ability of the system to retrieve relevant images for a given $n$-gram query. Specifically, we rank all images in the test set according to the calibrated log-likelihood our models predict for the query-image pairs.

In Figure 3, we show examples of twelve images that are most relevant from a set of $931,588$ YFCC100M test images (according to our model) for four different $n$-gram queries; we manually picked these $n$-grams to demonstrate the merits of building phrase-level image recognition models. The figure shows that the model has learned accurate visual representations for $n$-grams such as "Market Street" and "street market", as well as for "city park" and "Park City" (see the caption of Figure 3 for details on the queries). We show a second set of image retrieval examples in Figure 4, which shows that our model is able to distinguish visual concepts related to Washington: namely, between the state, the city, the baseball team, and the hockey team.

As in our earlier experiments, we quantify the image-retrieval quality of our model on a set of $10,000$ test images from the YFCC100M dataset by measuring the precision and recall of retrieving the correct image given a query $n$-grams. We compute a precision-recall curve by averaging over the $10,000$ $n$-gram queries that have the highest tf-idf value in the YFCC100M dataset: the resulting curve is shown in Figure 5. The results from this experiment are in accordance with the previous results: the naive $n$-gram loss substantially outperforms our Imagenet baseline, which in turn, is outperformed by the model trained using Jelinek-Mercer loss. Admittedly, the precisions we obtain are fairly low even in the low-recall regime. This low recall is the result of the false-negative noise in the "ground truth" we use for evaluation: an image that is relevant to the $n$-gram query may not be associated with that $n$-gram in the YFCC100M dataset, as a result of which we may consider it as "incorrect" even when it ought to be correct based on the visual content of the image.

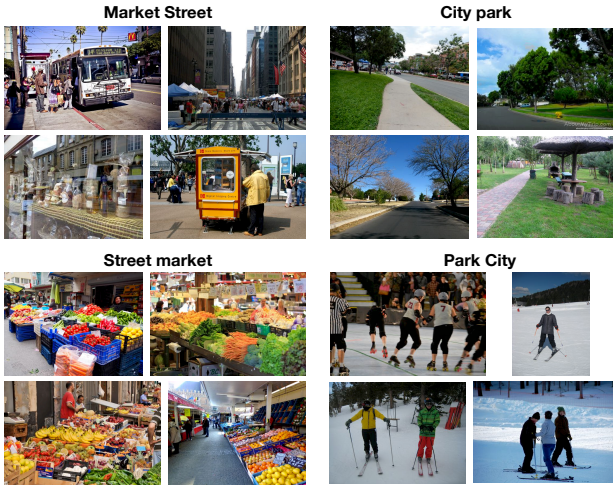

Figure 3. Four highest-scoring images for $n$-gram queries "Market Street", "street market", "city park", and "Park City" from a collection of $931,588$ YFCC100M images. Market Street is a common street name, for instance, it is one of the main thoroughfares in San Francisco. Park City (Utah) is a popular winter sport destination. The figure only shows images from the YFCC100M dataset whose license allows reproduction. We refer to the supplementary material for detailed copyright information.

### 4.3. Relating Images and Captions

In the third set of experiments, we study to whether visual $n$-gram models can be used for relating images and captions. While many image-conditioned language models have focused on caption generation, accurately measuring the quality of a model is still an open problem: most current metrics poor correlated with human judgement [1]. Therefore, we focus on caption-based retrieval tasks instead: in particular, we evaluate the performance of our models in caption-based image retrieval and image-based caption re-trieval. In caption-based image retrieval, we rank images according to their log-likelihood for a particular caption and measure recall@$k$: the percentage of queries for which the correct image is among the $k$ first images.

| Model | R@1 | R@5 | R@10 | Accuracy |
|---|---|---|---|---|
| Imagenet + linear | 1.1 | 3.3 | 4.8 | 38.3 |
| Naive $n$-gram | 1.3 | 4.4 | 6.9 | 42.0 |
| Jelinek-Mercer | **7.1** | **16.7** | **21.5** | **53.1** |

Table 3. Caption retrieval performance on YFCC100M test set of $10,000$ images measured in terms of recall@$k$ at three cut-off levels $k$ (lefthand-side; see text for details) and the percentage of correctly retrieved captions according to human raters (righthand-side) one baseline model and two of our phrase prediction models. Higher is better.

| Image retrieval | COCO-5K | | | Flickr-30K | | |
|---|---|---|---|---|---|---|
| | R@1 | R@5 | R@10 | R@1 | R@5 | R@10 |
| **Retrieval models** | | | | | | |
| Karpathy et al. [30] | – | – | – | 10.2 | 30.8 | 44.2 |
| Klein et al. [33] | **11.2** | **29.2** | **41.0** | 25.0 | 52.7 | 66.0 |
| Deep CCA [65] | – | – | – | 26.8 | 52.9 | 66.9 |
| Wang et al. [60] | – | – | – | **29.7** | **60.1** | **72.1** |
| **Language models** | | | | | | |
| STD-RNN [50] | – | – | – | 8.9 | 29.8 | 41.1 |
| BRNN [29] | 10.7 | 29.6 | 42.2 | 15.2 | 37.7 | 50.5 |
| Kiros et al. [32] | – | – | – | 16.8 | 42.0 | 56.5 |
| NIC [59] | – | – | – | 17.0 | – | 57.0 |
| **Ours** | | | | | | |
| Naive $n$-gram | 0.3 | 1.1 | 2.1 | 1.0 | 2.9 | 4.9 |
| Jelinek-Mercer | 5.0 | 14.5 | 21.9 | 8.8 | 21.2 | 29.9 |
| J-M + finetuning | 11.0 | 29.0 | 40.2 | 17.6 | 39.4 | 50.8 |

Table 4. Recall@$k$ (for three cut-off levels $k$) of caption-based image retrieval on the COCO-5K and Flickr-30K datasets for eight baseline models and our models (with and without finetuning). Baselines are separated in models dedicated to retrieval (top) and image-conditioned language models (bottom). Higher is better.

We first perform an experiment on $10,000$ images and comments from the YFCC100M test set. In addition to recall@$k$, we also measure accuracy by asking two human raters to assess whether the retrieved caption is relevant to the image content. The results of these experiments are presented in Table 3: they show that the strong performance of our visual $n$-gram models extends to caption retrieval[3]. According to human raters, our best model retrieves a relevant caption for $53.1\%$ of the images in the test set. To assess if visual $n$-grams help, we also experiment with a unigram model [28] with a dictionary size of $142,806$. We find that this model performs worse than visual $n$-gram models: its recall@$k$ scores of are $1.2$, $4.2$, and $6.3$, respectively.

---
[3]We also performed experiments with a neural image captioning model that was trained on COCO [59], but this model performs poorly: it obtains a recall@$k$ of 0.2, 1.0, and 1.6 for $k = 1$, 5, and 10, respectively. This is because many of the words that appear in YFCC100M are not in COCO.

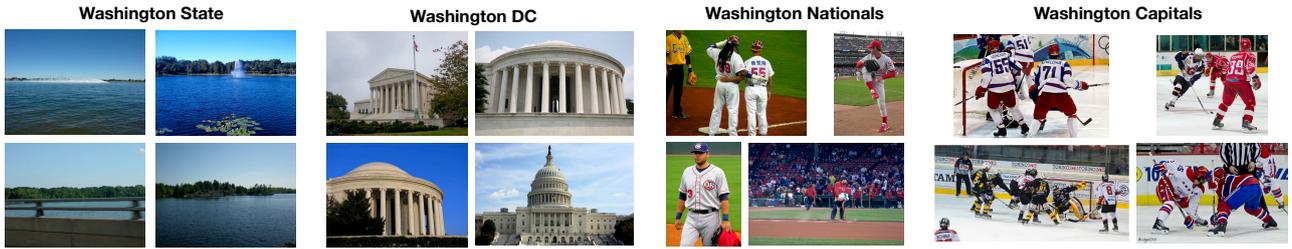

Figure 4. Four highest-scoring images for $n$-gram queries "Washington State", "Washington DC", "Washington Nationals", and "Washington Capitals" from a collection of $931,588$ YFCC100M test images. Washington Nationals is a Major League Baseball team; Washington Capitals is a National Hockey League hockey team. The figure only shows images from the YFCC100M dataset whose license allows reproduction. We refer to the supplementary material for detailed copyright information.

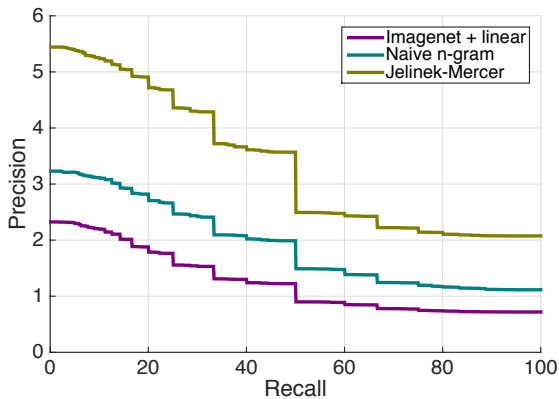

Figure 5. Precision-recall curve for phrase-based image retrieval of our models on YFCC100M test set of $10,000$ images one baseline model and two of our phrase-prediction models. The curves were obtained by averaging over the $10,000$ $n$-gram queries with the highest tf-idf value.

| **Caption retrieval** | **COCO-5K** | | | **Flickr-30K** | | |
|---|---|---|---|---|---|---|
| | R@1 | R@5 | R@10 | R@1 | R@5 | R@10 |
| **Retrieval models** | | | | | | |
| Karpathy *et al.* [30] | – | – | – | 16.4 | 40.2 | 54.7 |
| Klein *et al.* [33] | 17.7 | 40.1 | 51.9 | 35.0 | 62.0 | 73.8 |
| Deep CCA [65] | – | – | – | 27.9 | 56.9 | 68.2 |
| Wang *et al.* [60] | – | – | – | **40.3** | **68.9** | **79.9** |
| **Language models** | | | | | | |
| STD-RNN [50] | – | – | – | 9.6 | 29.8 | 41.1 |
| BRNN [29] | 16.5 | 39.2 | 52.0 | 22.2 | 48.2 | 61.4 |
| Kiros *et al.* [32] | – | – | – | 23.0 | 50.7 | 62.9 |
| NIC [59] | – | – | – | 23.0 | – | 63.0 |
| **Ours** | | | | | | |
| Naive $n$-gram | 0.7 | 2.8 | 4.6 | 1.2 | 5.9 | 9.6 |
| Jelinek-Mercer | 8.7 | 23.1 | 33.3 | 15.4 | 35.7 | 45.1 |
| J-M + finetuning | **17.8** | **41.9** | **53.9** | 28.6 | 54.7 | 66.0 |

Table 5. Recall@$k$ (for three cut-off levels $k$) of caption retrieval on the COCO-5K and Flickr-30K datasets for eight baseline systems and our visual $n$-gram models (with and without finetuning). Baselines are separated in models dedicated to retrieval (top) and image-conditioned language models (bottom). Higher is better.

To facilitate comparison with existing methods, we also perform experiments on the COCO-5K and Flickr-30K datasets [42, 66] using visual $n$-gram models trained on YFCC100M[4]. The results of these experiments are presented in Table 4; they show that our model performs roughly on par with the state-of-the-art based on language models on both datasets. We emphasize that our models have much larger vocabularies than the baseline models, which implies the strong performance of our models likely generalizes to a much larger visual vocabulary than the vocabulary required to perform well on COCO-5K and Flickr-30K. Like other language models, our models perform worse on the Flickr-30K dataset than dedicated retrieval models [30, 33, 60, 65]. Interestingly, our model does perform on par with a state-of-the-art retrieval model [33] on COCO-5K.

---
[4]Please see supplementary materials for additional results in COCO-1K and additional baseline models for relating images and captions.

We also perform image-based caption retrieval experiments: we retrieve captions by ranking all captions in the COCO-5K and Flick-30K test set according to their log-likelihood under our model. The results of this experiment are presented in Table 5, which shows that our model performs on par with state-of-the-art image-conditioned language models on caption retrieval. Like all other language models, our model performs worse than approaches tailored towards retrieval on the Flickr-30K dataset. On COCO-5K, visual $n$-grams perform on par with the state-of-the-art.

### 4.4. Zero-Shot Transfer

Because our models are trained on approximately 30 million photos and comments, they have learned to recognize a wide variety of visual concepts. To assess the ability of our models to recognize visual concepts out-of-the-box, we perform a series of *zero-shot transfer* experiments. Unlike traditional zero-shot learners (*e.g.*, [7, 35, 69]), we simply apply the Flickr-trained models on a test set from a different dataset. We automatically match the classes in the target dataset with the $n$-grams in our dictionary. We perform experiments on the aYahoo dataset [16], the SUN dataset [63], and the Imagenet dataset [10]. For a test image, we rank the classes that appear in each dataset according to the score our model assigns to the corresponding $n$-grams, and predict the highest-scoring class for that image. We report the accuracy of the resulting classifier in Table 6 in two settings: (1) a setting in which performance is measured only on in-dictionary classes and (2) a setting in which performance is measured on all classes.

|  | aYahoo | Imagenet | SUN |
| --- | --- | --- | --- |
| Class mode (in dictionary) | 15.3 | 0.3 | 13.0 |
| Class mode (all classes) | 12.5 | 0.1 | 8.6 |
| Jelinek-Mercer (in dictionary) | 88.9 | 35.2 | 34.7 |
| Jelinek-Mercer (all classes) | 72.4 | 11.5 | 23.0 |

Table 6. Classification accuracies on three zero-shot transfer learning datasets on in-dictionary and on all classes. The number of in-dictionary classes is 10 out of 12 for aYahoo, 326 out of 1,000 for Imagenet, and 330 out of 720 for SUN. Higher is better.

The results of these experiments are shown in Table 6. For reference, we also present the performance of a model that always predicts the a-priori most likely class. The results reveal that, even without any finetuning or re-calibration, non-trivial performances can be obtained on generic vision tasks. The performance of our models is particularly good on common classes such as those in the aYahoo dataset for which many examples are available in the YFCC100M dataset. The performance of our models is worse on datasets that involve fine-grained classification such as Imagenet, for instance, because YFCC100M contains few examples of specific, uncommon dog breeds.

## 5. Discussion and Future Work

**Visual $n$-grams and recurrent models.** This study has presented a simple yet viable alternative to the common practice of training a combination of convolutional and recurrent networks to relate images and language. Our visual $n$-gram models differ in several key aspects from models based on recurrent networks. Visual $n$-gram models are less suitable for caption generation[5] [44] but they are much more efficient to evaluate at inference time, which is very important in real-world applications of these models. Moreover, visual $n$-gram models can be combined with class activation mapping [48, 70] to perform visual grounding of $n$-grams, as shown in Figure 6. Such grounding is facilitated by the close relation between predicting visual $n$-grams and standard image classification. This makes visual $n$-gram models more amenable to transfer to new tasks than approaches based on recurrent models, as demonstrated by our zero-shot transfer experiments.

---
[5]Our model achieves a METEOR score [11] of 17.2 on COCO captioning with a test set of 1,000 images, versus 15.7 for a nearest neighbor baseline method and 19.5 for a recurrent network [29].

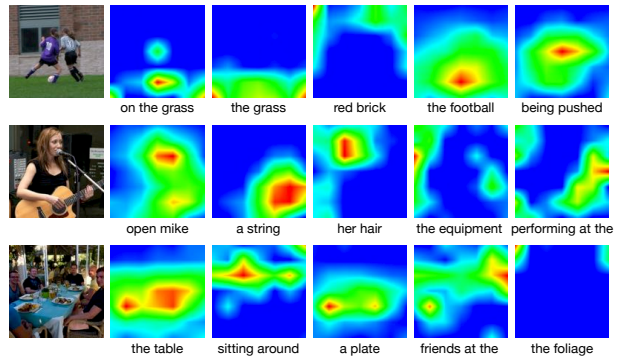

Figure 6. Discriminative regions of five $n$-grams for three images, computed using class activation mapping [48, 70].

**Learning from web data.** Another important aspect that discerns our work from most approaches in vision is that our models are capable of being learned purely from web data, *without any manual data annotation*. We believe that this type of training is essential if we want to construct models that are not limited to a small visual vocabulary and that are readily applicable to real-world computer-vision tasks. Indeed, this paper fits in a recent line of work [9, 28] that abandons the traditional approach of gathering images, manually annotating them for a small visual vocabulary, and training and testing on the resulting image-target distribution. As a result, models such as ours may not necessarily achieve state-of-the-art results on established benchmarks, because they did not learn to exploit the biases of those benchmarks as well [25, 51, 52]. Such "negative" results highlight the necessity of developing less biased benchmarks that provide more signal on progress towards visual understanding.

**Future work.** The Jelinek-Mercer loss we studied in this paper is based on just one of many $n$-gram smoothers [22]. In future work, we plan to perform an in-depth comparison of different smoothers for the training of convolutional networks. In particular, we will consider loss functions based as absolute-discounting smoothing such as Kneser-Ney smoothing [34], as well as back-off models [31]. We also plan to explore the use of visual $n$-gram models in systems that operate in open-world settings, combining them with techniques for zero-shot and few-shot learning. Finally, we aim to use our models in tasks that require recognition of a large variety of visual concepts and relations between them, such as visual question answering [2, 67], visual Turing tests [21], and scene graph prediction [36].

# Supplementary Material for
# "Learning Visual N-Grams from Web Data"

Ang Li[*]
University of Maryland
College Park, MD 20742, USA
angli@umiacs.umd.edu

Allan Jabri   Armand Joulin   Laurens van der Maaten
Facebook AI Research
770 Broadway, New York, NY 10025, USA
{ajabri,ajoulin,lvdmaaten}@fb.com

## 1. Introduction

The supplementary material for the submission "Learning Visual N-Grams for Web Data" is presented below. In Section 2, we provide all license information for all images from the YFCC100M dataset that were used in the main paper. In Section 3, we present quantitative results for image and caption retrieval on the COCO caption test set of $1,000$ images (COCO-1K). In Section 4, we present additional qualitative results of phrase prediction.

## 2. License Information for YFCC100M Photos

We reproduce all YFCC100M photos that appear in the main paper with relevant authorship and license information in Figure 1, 2, 3 and 4.

## 3. Relating Images and Captions: Additional Results

As an addition to the image and caption retrieval results on COCO-5K and Flickr-30K presented in the paper, we also provide retrieval results on the COCO-1K dataset, a test set of $1,000$ images provided by Karpathy and Fei-Fei [1]. In Table 1, we show the caption retrieval (left) and image retrieval (right) performance of four baseline models and our visual $n$-gram models on COCO-1K. We do not report results we obtained with the last version of the neural image captioning model [4] here because that model was trained on COCO validation set that was used as the basis for the COCO-1K test set.

The results on the COCO-1K dataset are in line with the results presented in the paper: our $n$-gram model performs roughly on par with recurrent language models [1, 3], but like these language models, it performs worse than models that were developed specifically for retrieval tasks [2, 5].

We provide additional results to demonstrate the effectiveness of end-to-end training. We trained a Jelinek-Mercer model on the ImageNet features as an additional

[*]This work was done while Ang Li was at Facebook AI Research.

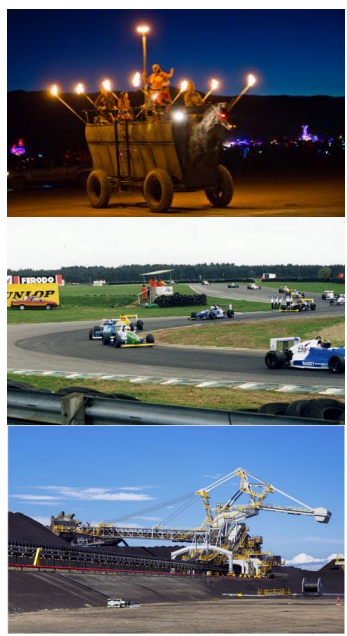

**Predicted $n$-grams**
lights
Burning Man
Mardi Gras
parade in progress

**Predicted $n$-grams**
GP
Silverstone Classic
Formula 1
race for the

**Predicted $n$-grams**
navy yard
construction on the
Port of San Diego
cargo

Figure 1. Four high-scoring visual $n$-grams for three images in our test set according to our visual $n$-gram model, which was trained *solely* on *unsupervised* web data. We selected the $n$-grams that are displayed in the figure from the five highest scoring $n$-grams according to our model, in such a way as to minimize word overlap between the $n$-grams. From top to bottom, photos are courtesy of: (1) Stuart L. Chambers (CC BY-NC 2.0); (2) Martin Pettitt (CC BY 2.0); (3) Gav Owen (C).

baseline and compare it with the end-to-end Jelinek-Mercer model in COCO-5K. The results are shown in Table 2 which reveals that an end-to-end trained Jelinek-Mercer model outperforms the one trained with ImageNet features in both non-finetuning and finetuning modes.

## 4. Phrase Prediction: Additional Results

We show additional qualitative results for predicting unigrams and bigrams in Figure 5; these examples were omit-

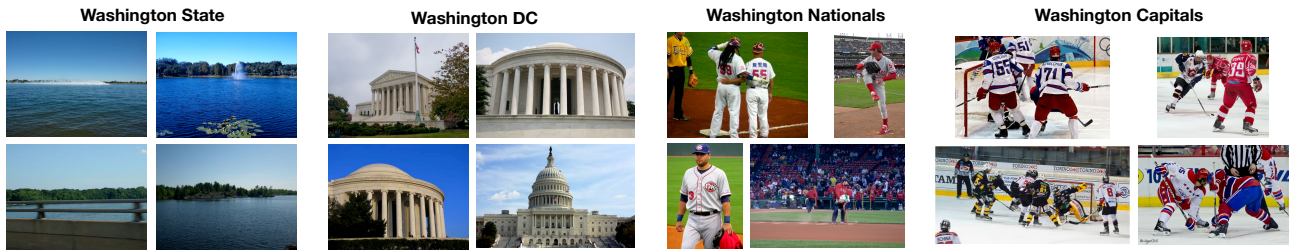

Figure 2. Four highest-scoring images for $n$-gram queries "Washington State", "Washington DC", "Washington Nationals", and "Washington Capitals" from a collection of 931,588 YFCC100M test images. Washington Nationals is a Major League Baseball team; Washington Capitals is a National Hockey League hockey team. The figure only shows images from the YFCC100M dataset whose license allows reproduction. From the top-left photo in clockwise direction, the photos are courtesy of: (1) Colleen Lane (CC BY-ND 2.0); (2) Ryaninc (CC BY 2.0); (3) William Warby (CC BY 2.0); (4) Cliff (CC BY 2.0); (5) Boomer-44 (CC BY 2.0); (6) Dannebrog (CC BY-ND 2.0); (7) S. Yume (CC BY 2.0); (8) Bridget Samuels (CC BY-NC-ND 2.0); (9) David G. Steadman (Public Domain Mark 1.0); (10) Hockey Club Torino Bulls (CC BY 2.0); (11) Brent Moore (CC BY-NC 2.0); (12) Andrew Malone (CC BY 2.0); (13) Terren in Virginia (CC BY 2.0); (14) Guru Sno Studios (CC BY-ND 2.0); (15) Derek Hatfield (CC BY 2.0); and (16) Bruno Kussler Marques (CC BY 2.0).

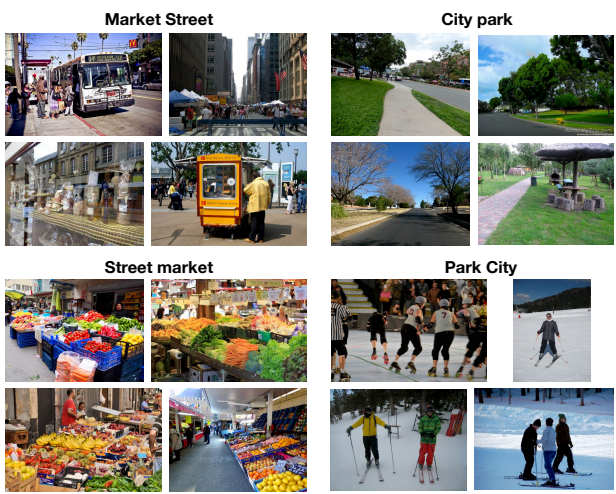

Figure 3. Four highest-scoring images for $n$-gram queries "Market Street", "street market", "city park", and "Park City" from a collection of 931,588 YFCC100M images. Market Street is a common street name, for instance, it is one of the main thoroughfares in San Francisco. Park City (Utah) is a popular winter sport destination. The figure only shows images from the YFCC100M dataset whose license allows reproduction. From left to right, photos are courtesy of the following photographers (license details between brackets). **Row 1:** (1) Jonathan Percy (CC BY-NC-SA 2.0); (2) Rachel Clarke (CC BY-NC-ND 2.0); (3) Richard Lazzara (CC BY-NC-ND 2.0); and (4) AboutMyTrip dotCom (CC BY 2.0). **Row 2:** (1) Alex Holyoake (CC BY 2.0); (2) Marnie Vaughan (CC BY-NC 2.0); (3) Hector E. Balcazar (CC BY-NC 2.0); and (4) Marcin Chady (CC BY 2.0). **Row 3:** (1) Rien Honnef (CC BY-NC-ND 2.0); (2) IvoBe (CC BY-NC 2.0); (3) Daniel Hartwig (CC BY 2.0); and (4) Benjamin Chodroff (CC BY-NC-ND 2.0). **Row 4:** (1) Guido Bramante (CC BY 2.0); (2) Alyson Hurt (CC BY-NC 2.0); (3) Xavier Damman (CC BY-NC-ND 2.0); and (4) Cassandra Turner (CC BY-NC 2.0).

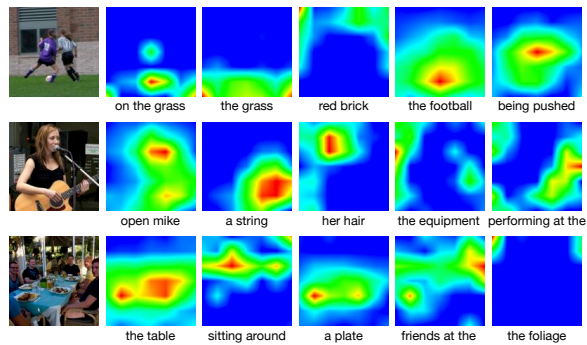

Figure 4. Discriminative regions of five $n$-grams for three images, computed using class activation mapping. From top to down, photos are courtesy of the following photographers (license details between brackets). **Row 1:** DebMomOf3 (CC BY-ND 2.0). **Row 2:** fling93 (CC BY-NC-SA 2.0). **Row 3:** Magnus (CC BY-SA 2.0).

| COCO-1K | Caption retrieval | | | Image retrieval | | |
|---|---|---|---|---|---|---|
| | R@1 | R@5 | R@10 | R@1 | R@5 | R@10 |
| **Retrieval models** | | | | | | |
| Klein *et al*. [2] | 38.9 | 68.4 | 80.1 | 25.6 | 60.4 | 76.8 |
| Wang *et al*. [5] | 50.1 | 79.7 | 89.2 | 39.6 | 75.2 | 86.9 |
| **Language models** | | | | | | |
| BRNN [1] | 38.4 | 69.9 | 80.5 | 27.4 | 60.2 | 74.8 |
| M-RNN [3] | 41.0 | 73.0 | 83.5 | 29.0 | 42.2 | 77.0 |
| **Ours** | | | | | | |
| Naive $n$-gram | 3.1 | 9.2 | 14.6 | 1.1 | 4.2 | 7.3 |
| Jelinek-Mercer | 22.5 | 47.6 | 60.7 | 12.8 | 33.5 | 46.5 |
| J-M + finetuning | 39.9 | 70.5 | 82.5 | 25.4 | 55.8 | 70.2 |

Table 1. Recall@$k$ (for three cut-off levels $k$) of caption and image retrieval on the COCO-1K dataset for three baseline systems and our visual $n$-gram models (with and without finetuning). Baselines are separated in models dedicated to retrieval (top) and image-conditioned language models (bottom). Higher is better.

| COCO-5K | Caption retrieval | | | Image retrieval | | |
|---|---|---|---|---|---|---|
| | R@1 | R@5 | R@10 | R@1 | R@5 | R@10 |
| Imagenet + J-M | 8.0 | 21.6 | 31.2 | 4.4 | 14.0 | 21.5 |
| End-to-end J-M | 8.7 | 23.1 | 33.3 | 5.0 | 14.5 | 21.9 |
| Imagenet + J-M (finetuning) | 12.7 | 31.0 | 43.0 | 6.5 | 18.9 | 28.1 |
| End-to-end J-M (finetuning) | 17.8 | 41.9 | 53.9 | 11.0 | 29.0 | 40.2 |

Table 2. Recall@$k$ (for three cut-off levels $k$) of caption and image retrieval on the COCO-5K dataset for four variants of our visual $n$-gram models (with and without finetuning). Higher is better.

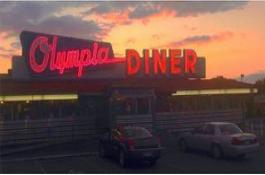

| **Unigrams** | **Bigrams** |
|---|---|
| Sign | Neon sign |
| Bar | Motel in |
| Ave | Store in |
| Store | Sign for |
| Diner | Sacramento CA |

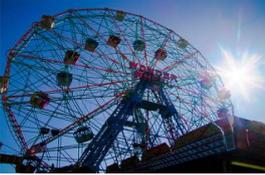

| Ferris | Ferris wheel |
| Blue | Lafayette Park |
| Wheel | Coney Island |
| Lafayette | Blue sky |
| Tower | Amusement park |

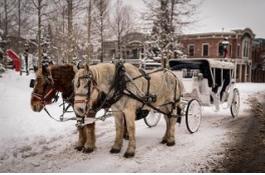

| Carriage | Horse drawn |
| Winter | Horse and |
| Horse | Winter in |
| Snow | Blizzard of |
| Blizzard | Snowy day |

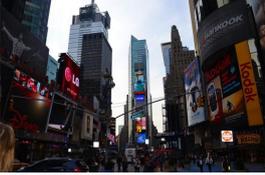

| Times | Times Square |
| Shinjuku | Shinjuku Tokyo |
| Ginza | Manhattan new |
| Manhattan | Hong Kong |
| NYC | Eaton Center |

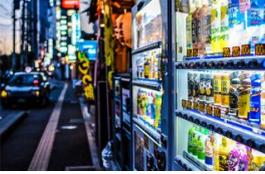

| Tokyo | Shinjuku Tokyo |
| Osaka | Tokyo Japan |
| Shinjuku | Vending machine |
| Vending | Osaka Japan |
| Store | Store in |

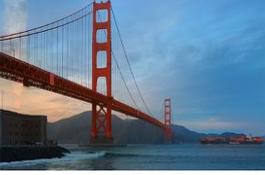

| Golden | Golden Gate |
| Marin | Suspension bridge |
| Suspension | Mackinac Island |
| Cruise | Oracle Team |
| Forth | Brooklyn Bridge |

Figure 5. Five highest-scoring visual unigrams and bigrams for five images in our test set. From top to bottom, photos are courtesy of: (1) Mike Mozart (CC BY 2.0); (2) owlpacino (CC BY-ND 2.0); (3) brando.n (CC BY 2.0); (4) Laura (CC BY-NC 2.0); (5) inefekt69 (CC BY-NC-ND 2.0); and (6) Yahui Ming (CC BY-NC-ND 2.0).

ted from the main paper because of space limitations.